\documentclass{article}
\usepackage{spconf,amsmath,graphicx}
\usepackage{amssymb}
\usepackage{xspace}
\usepackage{tikz}
\usetikzlibrary{arrows.meta, positioning, shapes.geometric}

\usepackage{graphicx}
\usepackage{booktabs}

\newcommand{\ourmethod}{$\mathcal{A}$rtContext \xspace}

\title{$\mathcal{A}$\MakeLowercase{rt}C\MakeLowercase{ontext}: Contextualizing Artworks with Open-Access Art History Articles and Wikidata Knowledge through a LoRA-Tuned CLIP Model}
\name{Samuel Waugh and Stuart James}
\address{Department of Computer Science\\ Durham University\\Durham, UK}
\begin{document}
%\ninept
%
\maketitle
\begin{abstract}
Many Art History articles discuss artworks in general as well as specific parts of works, such as layout, iconography, or material culture. However, when viewing an artwork, it is not trivial to identify what different articles have said about the piece. Therefore, we propose \ourmethod, a pipeline for taking a corpus of Open-Access Art History articles and Wikidata Knowledge and annotating Artworks with this information. We do this using a novel corpus collection pipeline, then learn a bespoke CLIP model adapted using Low-Rank Adaptation (LoRA) to make it domain-specific. We show that the new model, PaintingCLIP, which is weakly supervised by the collected corpus, outperforms CLIP and provides context for a given artwork. The proposed pipeline is generalisable and can be readily applied to numerous humanities areas. 
\end{abstract}
\begin{figure}[t!]
    \centering
    \includegraphics[width=1.0\linewidth]{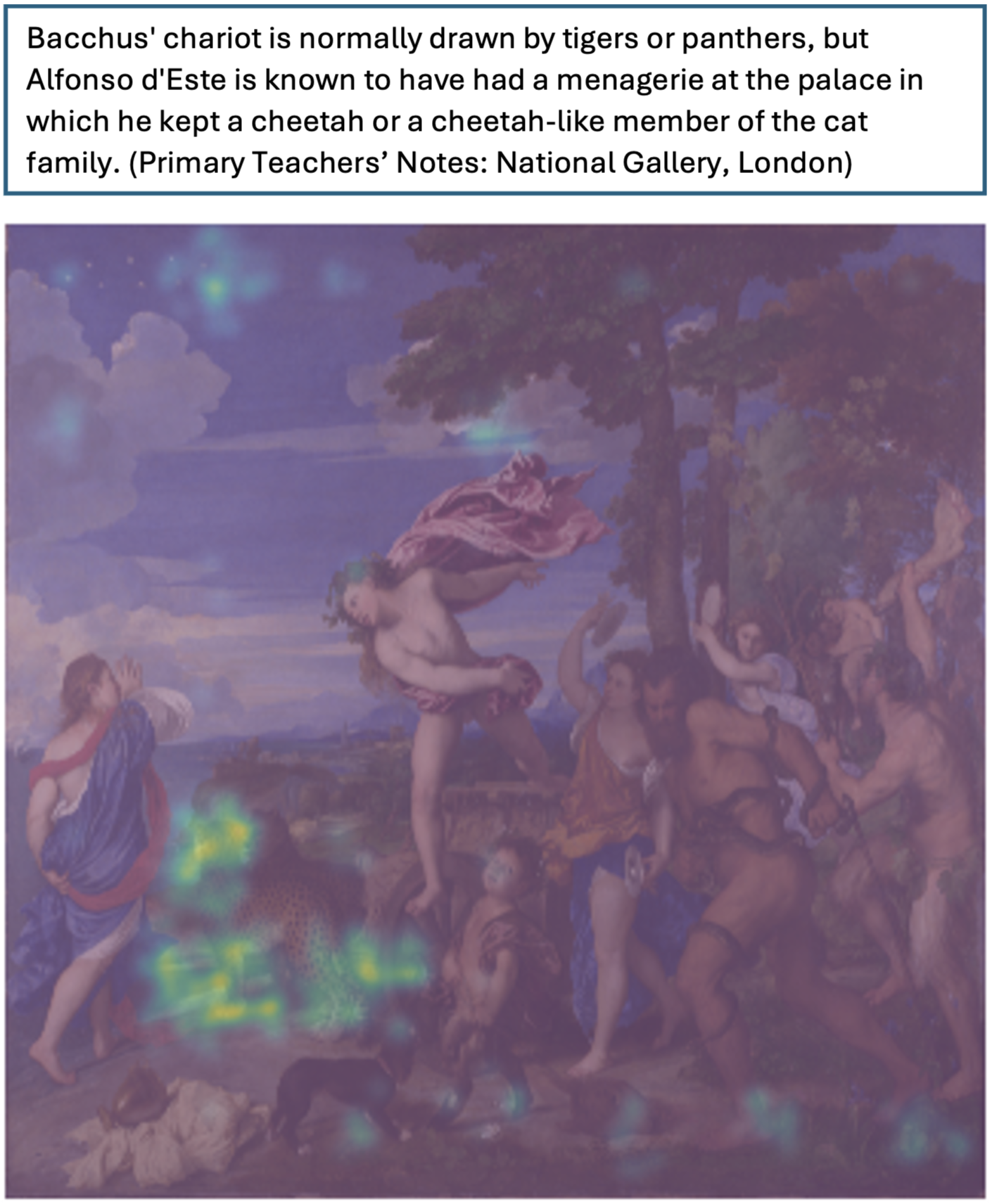} 
    \caption{\textit{Bacchus and Ariadne} (Titian, 1523) \,--\, CLIP-based saliency visualization of a Heat-map overlay produced by back-propagating image–text similarity for the sentence; warmer regions mark the most influence on the model’s score.}
    \label{fig:bacchus-heatmap}
\end{figure}
\section{Introduction}
\label{sec:intro}

When viewers encounter a painting, the extensive art-historical scholarship that explains its iconography, technique, and cultural meaning remains largely inaccessible. Bridging this gap is a central challenge at the intersection of computer vision, natural-language processing, and digital humanities. Grounding painting descriptions in authoritative labels drawn from the art-history literature would enrich public engagement while advancing domain-adapted vision–language modeling.

Existing work largely treats paintings as a visual labeling problem \cite{van2015toward}, training convolutional networks to classify author or style \cite{saleh2015unified}. While useful for indexing or stylistic generation, such approaches do not explain what a painting depicts or how its imagery has been interpreted by scholars. As with human viewers, images alone are often insufficient to convey artistic intent without contextual text \cite{serrell2015exhibit}. Even object-level annotations fail to capture meaning. A painting such as Bacchus and Ariadne (Titian, 1523) cannot be understood by enumerating detected figures: Bacchus’ leap signifies a pivotal narrative moment. Capturing such symbolism requires linking visual elements to the textual scholarship that explains their significance—an ability absent from current computer-vision pipelines.

Vision-language models, such as CLIP \cite{radford2021learning}, embed images and full sentences in a shared space, providing a promising foundation for this task. To adapt CLIP to art history without costly retraining and extensive strong supervision data, we employ Low-Rank Adaptation (LoRA) \cite{hu2022lora}, which learns a small set of task-specific parameters while preserving the model’s general capabilities. This lightweight adaptation is essential given the limited scale of openly licensed art-historical text compared to domains such as biomedicine, where specialized language models like SciBERT \cite{beltagy2019scibert} are feasible with established large-scale training datasets.

To address the training problem, we combine open-access art-history literature with structured metadata from Wikidata \cite{vrandevcic2014wikidata}, which provides factual context often implicit in scholarly writing. Although much art-historical scholarship remains closed, sufficient open material exists. Therefore, we curated a corpus of open-access art-history articles and then built an end-to-end pipeline that weakly aligns paintings with scholarly sentences and fine-tunes CLIP using LoRA. Our pipeline, \ourmethod, harvests articles via OpenAlex, extracts and segments text, aligns sentences to artworks using Sentence-BERT \cite{reimers2019sentence} and Wikidata metadata, and uses these pairs to fine-tune CLIP ViT-B/32. The resulting model, PaintingCLIP, is evaluated using averaged precision–recall and qualitative error analysis.

Our contributions are therefore: 
\textit{1)} a large-scale ingestion pipeline for creating corpora, in our case, 27,044 open-access art-history PDFs across 450 artists; \textit{2)} a weakly knowledge graph supervised training extraction pipeline extracting key sentences and relations between visual and textual content for training; and \textit{3)} a LoRA adaptation for domain-specific matching of the CLIP model called PaintingCLIP. 

\section{Related Work}
We situate \ourmethod with respect to prior work on automatic art analysis, vision--language models, lightweight domain adaptation, and the use of scholarly text as supervision. 

\begin{figure*}
    \centering
    \includegraphics[width=\textwidth]{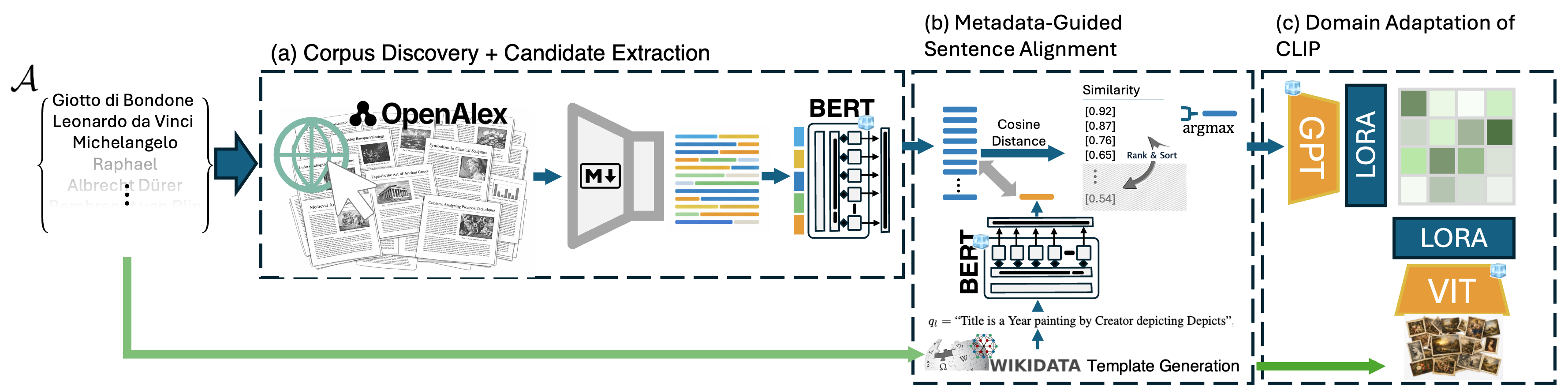}
    \caption{Overview of \ourmethod. From a set of artist $\mathcal{A}$, (a) Open-access art-historical articles are harvested via OpenAlex and converted into structured text, from which candidate sentence contexts are extracted and embedded using Sentence-BERT. (b) For each painting, Wikidata metadata is used to construct a semantic query that is matched against candidate sentences to select the most relevant description. (c) The resulting image–text pairs supervise Low-Rank Adaptation (LoRA) fine-tuning of CLIP ViT-B/32, producing PaintingCLIP with improved alignment between paintings and art-historical articles.}

    \label{fig:ds}
\end{figure*}

\noindent\textbf{Automatic Art Analysis.} Early work moved beyond style classification by aligning paintings with curatorial text. SemArt \cite{garcia2018read} framed art understanding as a cross-modal retrieval problem, embedding images and descriptions in a shared space. Context was later incorporated implicitly through multi-task learning or explicitly via knowledge graphs \cite{garcia2019context}. While these approaches improved semantic alignment, they relied on image representations and supervised labels. Richer metadata followed. ArtGraph \cite{castellano2022leveraging} combined visual features with structured knowledge from WikiArt and DBpedia via graph-based representations. Despite classification gains, such methods remain image-level and do not ground outputs in scholarly prose.

Captioning work has sought to encode symbolic meaning more directly. Cetinic \cite{cetinic2021towards} generated iconography-aware descriptions using Iconclass annotations \cite{cetinic2021iconographic}, but outputs were brief and lacked fine-grained attribution. Related efforts addressed object detection in art via style-augmented photographic data \cite{kadish2021improving}, improving robustness while inheriting dataset bias. More recently, large vision--language models have been evaluated on art tasks in zero-shot settings \cite{strafforello2024have}. Persistent errors on nuanced styles highlight the limits of general-purpose training without domain-specific grounding.\\

\noindent\textbf{Vision--Language Models \& Domain Adaptation.}
CLIP \cite{radford2021learning} aligns images and text through large-scale contrastive training, enabling zero-shot recognition via natural language prompts. Applied directly to art datasets, it outperforms traditional CNNs \cite{baldrati2022exploiting} but remains sensitive to domain mismatch. Extensions address these limitations through region-level alignment \cite{zhong2022regionclip}. While effective for object-centric tasks, these methods do not explicitly connect artworks to expert-written scholarship.

Efficient domain adaptation methods avoid full fine-tuning. Adapters and prompt-based approaches \cite{houlsby2019parameter,lester2021power} modify frozen models with task-specific parameters. LoRA \cite{hu2022lora} instead injects low-rank updates into existing weights, enabling efficient adaptation with minimal overhead, making it suitable for domains with limited labeled data.\\

\noindent\textbf{Scholarly Text as Supervision.} Text summarisation methods are commonly divided into abstractive and extractive approaches \cite{sharma2022automatic}. Abstractive models based on large transformers \cite{devlin2019bert} generate fluent summaries but risk factual drift, particularly in specialized domains \cite{nechakhin2024evaluating}. Extractive methods preserve original phrasing and terminology, making them more appropriate for scholarly text. Modern extractive techniques rely on sentence embeddings and structural cues \cite{nallapati2017summarunner}. Sentence-BERT \cite{reimers2019sentence} enables efficient semantic selection of relevant sentences from scholarly corpora. Despite their effectiveness, such NLP pipelines are rarely integrated with visual models.

\section{Methodology}
\label{sec:method}
We address the problem of grounding paintings in expert-written art-historical articles by constructing weak supervision between images and scholarly text, and using this supervision to adapt a pre-trained vision–language model. Let $\mathcal{I}$ denote a set of digital images of paintings and $\mathcal{T}$ a corpus of art-historical text. The objective is to learn a joint embedding function $f : \mathcal{I} \cup \mathcal{T} \rightarrow \mathbb{R}^d$, such that each painting image is close, in embedding space, to textual descriptions that interpret its visual content. Explicit image–text alignments are unavailable; instead, we construct approximate alignments using structured metadata and semantic similarity.

Our approach, \ourmethod, consists of four stages: (i) corpus discovery, (ii) candidate context extraction, (iii) metadata-guided sentence alignment, and (iv) domain adaptation of CLIP. Together, these stages define a mapping from a large, weakly structured text corpus to a set of image–text training pairs suitable for contrastive learning.

\subsection{Corpus Discovery} \label{subsec:discovery}

The first stage constructs a corpus of open-access art-historical articles from which candidate descriptions can be extracted. We use OpenAlex \cite{priem2022openalex}, which indexes a universe of works $\mathcal{W}$, each annotated with a set of topic tags drawn from a controlled vocabulary $\mathcal{T}_{\text{all}}$. For each work $\omega \in \mathcal{W}$, we extract all tags $\operatorname{tags}(\omega) \subseteq \mathcal{T}_{\text{all}}$.

To restrict the corpus to art-historical material, we define a subset $\mathcal{T}_{\text{art}} \subset \mathcal{T}_{\text{all}}$ consisting of $14$ topics drawn from the Art History subfield of the Social Sciences domain. A work is retained if $\operatorname{tags}(\omega) \cap \mathcal{T}_{\text{art}} \neq \varnothing$. 

Therefore, given an identified set of $450$ artists, let $\mathcal{A} = \{A_1,\dots,A_{n}\}$ denote the set of artists spanning multiple periods, movements, and cultural contexts manually curated, where $n = 450$. For each artist $A_i$, an OpenAlex query returns a ranked list of works $A_i = (A_{i1}, A_{i2}, \dots, A_{in}$, ordered by OpenAlex relevance score $r$, with $r > \rho$, where $\rho = 1.0$ is a fixed threshold. Queries are further constrained to English-language, open-access PDFs. This process yields a corpus of $27,044$ articles grouped by artist identity ($A_{ik}$).

\subsection{Candidate Context Extraction}\label{subsec:extraction}
Each artist article $A_{ik}$ is provided as a PDF, which we transform into a set of semantically meaningful textual contexts suitable for matching against paintings. We define an extraction process $\Phi : A_{ik} \mapsto S_{ik}$, implemented as a composition of three operators: \textit{(i)} conversion to structured Markdown; \textit{(ii)} context extraction and cleaning to obtain $C_{ik}$; and \textit{(iii)} representation generation to obtain embeddings $S_{ik}$.

First, we convert each PDF into structured Markdown text $M_{ik}$. To ensure tractable processing, only files smaller than $10\,\mathrm{MB}$ are retained, excluding long monographs that would otherwise create an artist and content bias.

We then map $M_{ik}$ to a set of candidate textual contexts (strings) $C_{ik}$. All non-textual elements are removed, and the remaining text is segmented into sentences using NLTK’s Punkt tokenizer. Sentences shorter than four tokens are discarded. For each remaining sentence, a context window is constructed by concatenating the preceding and following sentences when available, yielding short paragraphs that retain local semantic coherence.

Finally, we embed each context $c \in C_{ik}$ using Sentence-BERT \cite{reimers2019sentence}. Let $\phi:\mathcal{T}\rightarrow\mathbb{R}^{d_s}$ denote this embedding function. The resulting set of context embeddings is
\[
S_{ik} = \{ \phi(c) \mid c \in C_{ik} \} \subset \mathbb{R}^{d_s}.
\]
Sentence-BERT is chosen for its ability to embed multi-sentence text efficiently while preserving semantic similarity.

For each artist $A_i$, the aggregated context-embedding set is
\[
S_i = \bigcup_{k} S_{ik}.
\]

\subsection{Metadata-Guided Sentence Alignment}
To align each painting with the most relevant candidate context. We utilize templates extracted from Wikidata to contextualize a given painting.  Let \(P = \{\, p_{1},\, p_{2},\, \dots,\, p_{37\,449} \}\) be the set of paintings also in Wikidata. For each painting $p_l$, Wikidata provides structured fields including title, creator, year, movement, and depicted entities. From this metadata, we construct a natural-language query from a template:
\[
\begin{aligned}
q_l ={}& \text{``[Title] is a [Year] painting}\\
       & \text{by [Creator] depicting [Depicts]''}
\end{aligned}
\]
creating, for example, ``Café Terrace at Night is a 1888 painting by Vincent van Gogh depicting platform, gas burner, La Cité, lamp, sett, Arles, coffeehouse, chair, table, tree, night, sky, star, human''. We then embed the template sentence using the same Sentence-BERT model, yielding $w_l \in \mathbb{R}^{d_s}$. Restricting attention to contexts associated with the painting’s creator, we define the aligned sentence as
\[
s_l^{*} = \arg\max_{s \in S_i} \; \psi(s, w_l),
\]
where $\psi$ denotes cosine similarity and $S_i$ is the candidate set for artist $A_i^{\text{Name}}$. Artist-level grouping substantially reduces computational cost while reflecting the empirical observation that scholarly discussion of a painting almost always occurs within texts addressing its creator. The resulting alignment produces one sentence per painting, yielding $29,697$ aligned image–text pairs.

\subsection{PaintingCLIP: Domain Adaptation of CLIP}

To weakly supervise the training of CLIP \cite{radford2021learning}. For each painting $p_l$, we define a structured label
\[
p_l^{\text{Label}} = p_l^{\text{Wiki}} + s_l^{*},
\]
where Wikidata metadata provides factual grounding and $s_l^{*}$ supplies interpretive context. We denote by the same symbol $s_l^{*}$ the textual sentence whose Sentence-BERT embedding attains this maximum.
The text is truncated using CLIP’s tokenization function $\tau_{77}$.

Training data are therefore
\[
L = \{ (p_l^{\text{Image}}, \tau_{77}(p_l^{\text{Label}})) \}_{l=1}^{29\,697}.
\]

Rather than full fine-tuning, we apply Low-Rank Adaptation (LoRA) \cite{hu2022lora} to CLIP’s visual and text projection heads. Let $W$ denote a frozen projection matrix; LoRA parameterize updates as
\[
W' = W + \Delta W, \qquad \Delta W = BA,
\]
where $A \in \mathbb{R}^{r \times d}$, $B \in \mathbb{R}^{d \times r}$, and $r \ll d$. This yields efficient adaptation with negligible inference-time overhead. We use rank $r=16$, scaling $\alpha=32$, and dropout $0.05$.

Optimization follows CLIP’s contrastive objective. The resulting model, PaintingCLIP, retains CLIP’s general visual–language structure while embedding paintings closer to scholarly descriptions that interpret their visual content.

\section{Evaluation}

We evaluate \ourmethod at three complementary levels: 
(i) statistical analysis of the constructed corpus and supervision signal, 
(ii) quantitative comparison between CLIP and PaintingCLIP on a controlled retrieval task, and 
(iii) qualitative analysis of retrieved art-historical article sentences. 
Together, these evaluations assess both the feasibility of constructing weak supervision at scale and the extent to which domain adaptation improves semantic alignment.

\subsection{Corpus Statistics and Supervision Characteristics}
Applying the ingestion pipeline described in Section~\ref{subsec:discovery} \& ~\ref{subsec:extraction}, we construct a corpus 
$\mathcal{A}=\{A_1,\dots,A_{450}\}$ of open-access art-historical articles grouped by artist.
The corpus contains $27{,}044$ articles covering $450$ artists, with article relevance scores retained from OpenAlex metadata.

The distribution of articles per artist is highly skewed.
Let $|A_i|$ denote the number of articles associated with artist $i$.
Across the corpus, $|A_i|$ spans more than two orders of magnitude.
Well-studied artists such as Henri Matisse are associated with several hundred articles, while many artists have fewer than five.
We found that the majority of artists have fewer than $50$ relevant articles, reflecting the uneven coverage of the art-historical canon in open-access literature.
A complementary view of this imbalance is obtained by analyzing painting-level prominence.
For each painting $p\in P$, we use its Wikidata link count $p^{\text{LinkCount}}$ as a proxy for scholarly visibility. The resulting heavy-tailed distribution indicates that a small subset of canonical works dominates scholarly references, while the long tail receives sparse attention.
This skew directly impacts the availability and quality of weak supervision for fine-tuning.

After matching articles, paintings, and extracted sentences, we obtain a fine-tuning set
$L=\{t_1,\dots,t_{29{,}697}\}$ of image–text pairs.
Unmatched or malformed records account for approximately $20\%$ of initially indexed paintings, highlighting the practical challenges of large-scale alignment across heterogeneous sources.

\subsection{Quantitative Retrieval Evaluation}
We evaluate retrieval performance using a controlled painting-to-sentence ranking task.
Let $\Pi=\{\pi_1,\dots,\pi_{10}\}$ denote the ten paintings with the highest Wikidata link counts.
For each painting $\pi_i$ and model $M\in\{\text{CLIP},\text{PaintingCLIP}\}$,
we retrieve a ranked list of candidate sentences
$(x_{ij}, s_{ij})_{j=1}^{10}$,
where $s_{ij}=M(\pi_i,x_{ij})$.

Each sentence is assigned a binary relevance label $y_{ij}\in\{0,1\}$ based on factual correctness and coherence with respect to $\pi_i$.
This yields score vectors $\mathbf{s}^{(i)}$ and label vectors $\mathbf{y}^{(i)}$.
From these, we compute precision–recall (PR) curves using thresholding over similarity scores.
Following \cite{davis2006relationship}, curves are interpolated and upper-enveloped on a shared recall grid
$\mathcal{R}=\{0,0.01,\dots,1\}$.
The macro-averaged precision is then
\begin{equation}
\bar P(r)=\frac{1}{|\Pi|}\sum_{i=1}^{|\Pi|} P_i(r).
\label{eq:macroPR_eval}
\end{equation}

\begin{figure}[!t]
  \centering
  \includegraphics[width=1.0\columnwidth]{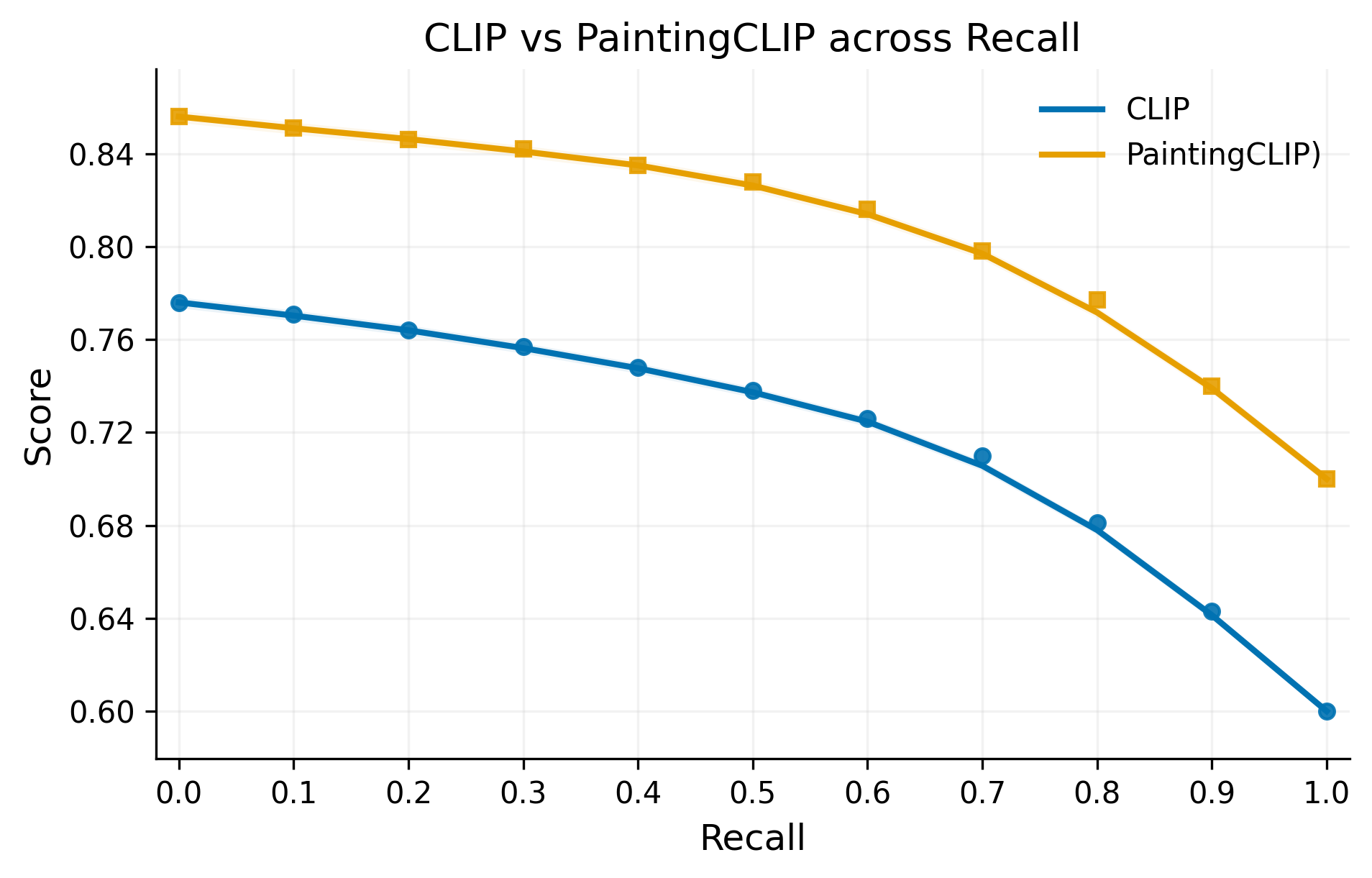}
  \caption{Averaged $\bar P(R)$ curves comparison over a set of queries for CLIP \textit{vs.} PaintingCLIP}
  \label{fig:precision_recall_graph}
\end{figure}

Fig.~\ref{fig:precision_recall_graph} shows that PaintingCLIP consistently dominates CLIP across recall levels.
The improvement is most pronounced in the high-precision regime, indicating that domain adaptation increases the likelihood that top-ranked sentences are genuinely descriptive.
While the evaluation set is necessarily small, this result provides quantitative evidence that weak supervision from art-historical articles improves semantic alignment.

\subsection{Qualitative Analysis}

\begin{figure}[!t]
  \centering
  \includegraphics[width=\linewidth]{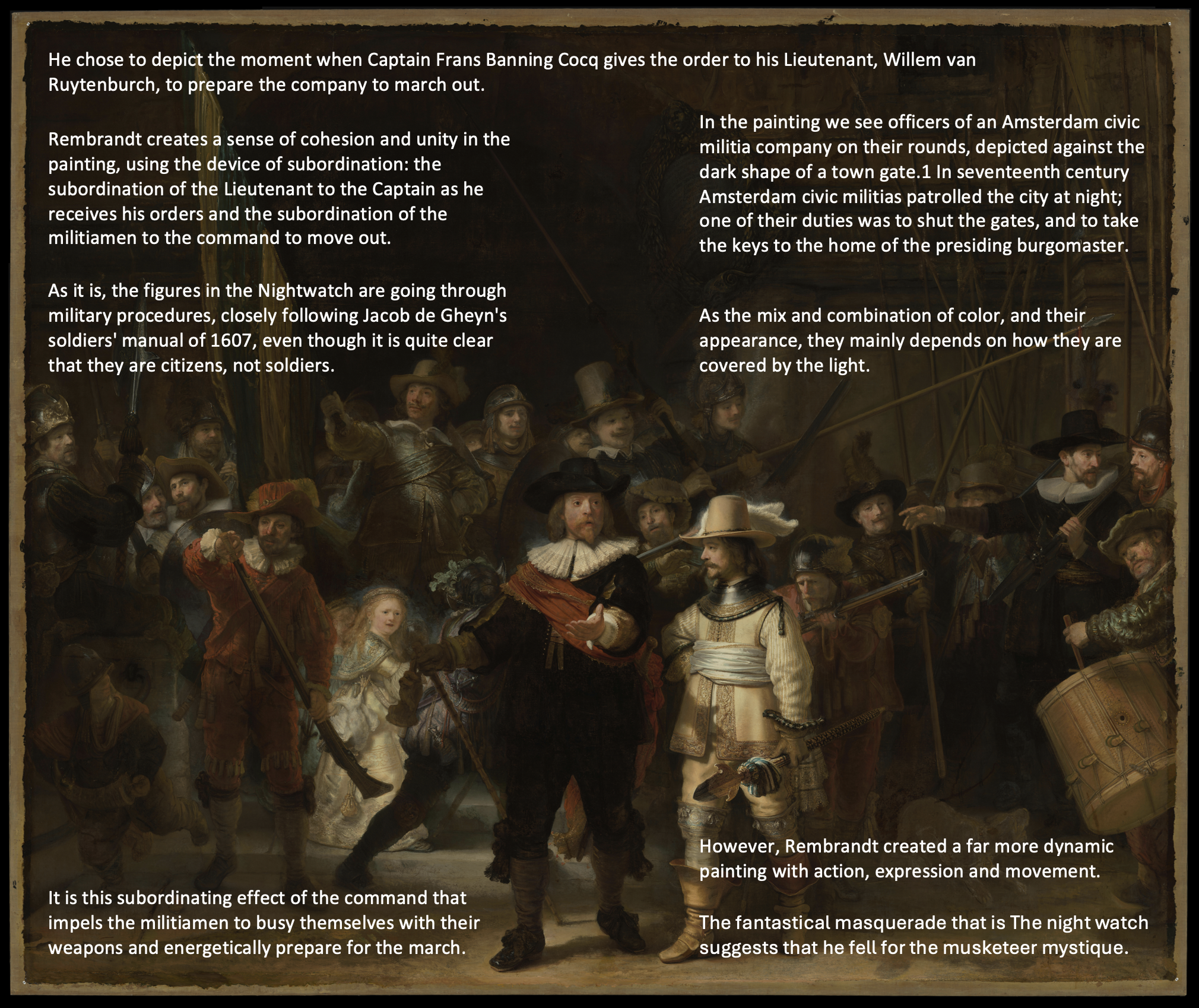}
  \caption{Example extracted sentences for \textit{The Night Watch} (Rembrandt van Rijn, 1642) using PaintingCLIP model and the corpus.}
  \label{fig:nightwatch}
\end{figure}

Quantitative improvements are reflected in qualitative retrieval behavior.
Fig.~\ref{fig:nightwatch} presents the sentences retrieved for \textit{The Night Watch} (Rembrandt van Rijn, 1642).
PaintingCLIP retrieves multiple coherent descriptions referring to composition, figures, and historical interpretation, whereas baseline CLIP frequently surfaces generic or weakly related statements.

\subsection{Limitations and Discussion}
Several limitations affect the current approach and evaluation.
The most common failure modes arise during sentence selection and include contexts that mention a painting only incidentally, fragmented text introduced by PDF parsing, and statements that are broadly plausible but require expert verification.
Notably, these errors originate primarily from noise and structure in the source articles rather than from visual misinterpretation, indicating that sentence selection quality remains a key challenge in the pipeline; nevertheless, the observed performance gains suggest that the approach generalizes despite this noise.

More broadly, the available supervision is biased toward canonical artists and well-documented works, reflecting structural imbalances in open-access art-historical publishing.
This skew propagates to the fine-tuning data and may limit performance on under-represented artists and movements.
Sentence selection currently relies on SBERT embeddings, which prioritizes coarse semantic similarity and may overlook finer iconographic or historiographic distinctions.
More expressive cross-encoder models or domain-adapted language representations could improve label quality, albeit at increased computational cost.

Finally, both the adaptation strategy and base model impose constraints.
LoRA is applied only to CLIP’s projection layers, without exploration of deeper adapter placement or alternative parameter-efficient tuning strategies.
In addition, CLIP operates on fixed $224\times224$ image crops and truncates text to $77$ tokens, which is misaligned with the fine-grained visual detail and extended descriptions typical of art-historical articles.

\section{Conclusion}
This paper examined whether a general-purpose vision--language model can be adapted, with limited supervision and modest computational cost, to associate paintings with descriptions drawn from art-historical articles. We introduced \ourmethod, a pipeline that constructs weak image--text supervision using structured metadata and semantic similarity, and applies parameter-efficient Low-Rank Adaptation to CLIP. The resulting model, PaintingCLIP, improves alignment between paintings and scholarly descriptions while preserving the base model’s zero-shot capabilities.

We constructed a corpus of 27\,044 open-access art-historical articles spanning 450 artists and derived 29\,697 image--text pairs for fine-tuning. Quantitative retrieval experiments on canonical works show consistent precision--recall improvements over the CLIP baseline, supported by qualitative evidence that the adapted model retrieves more specific and coherent descriptions.

Overall, the results demonstrate that weak supervision extracted from art-historical articles can be operationalized at scale to adapt vision--language models without full retraining, providing a practical route toward more semantically grounded representations for artworks.

\vfill\pagebreak

\bibliographystyle{IEEEbib}
\bibliography{refs}

\end{document}